\documentclass[sigconf]{acmart}
\usepackage{amsfonts}
\usepackage{subcaption}
\usepackage{csquotes}

\pdfoutput=1

\title[Detecting Environmental Violations with Satellite Imagery]{Detecting Environmental Violations with Satellite Imagery in Near Real Time: Land Application under the Clean Water Act}

\copyrightyear{2022}
\acmYear{2022}
\setcopyright{acmlicensed}\acmConference[CIKM '22]{Proceedings of the 31st ACM International Conference on Information and Knowledge Management}{October 17--21, 2022}{Atlanta, GA, USA}
\acmBooktitle{Proceedings of the 31st ACM International Conference on Information and Knowledge Management (CIKM '22), October 17--21, 2022, Atlanta, GA, USA}
\acmPrice{15.00}
\acmDOI{10.1145/3511808.3557104}
\acmISBN{978-1-4503-9236-5/22/10}



\begin{abstract}
   This paper introduces a new, highly consequential setting for the use of computer vision for environmental sustainability. Concentrated Animal Feeding Operations (CAFOs) (aka intensive livestock farms or ``factory farms'') produce significant manure and pollution. Dumping manure in the winter months poses significant environmental risks and violates environmental law in many states. Yet the federal Environmental Protection Agency (EPA) and state agencies have relied primarily on self-reporting to monitor such instances of ``land application.''  Our paper makes four contributions. First, we introduce the environmental, policy, and agricultural setting of CAFOs and land application. Second, we provide a new dataset of high-cadence (daily to weekly) 3m/pixel satellite imagery from 2018-20 for 330 CAFOs in Wisconsin with hand labeled instances of land application (n=57,697). Third, we develop an object detection model to predict land application and a system to perform inference in near real-time. We show that this system effectively appears to detect land application (PR AUC = 0.93) and we uncover several outlier facilities which appear to apply regularly and excessively. Last, we estimate the population prevalence of land application events in Winter 2021/22. We show that the prevalence of land application is much higher than what is self-reported by facilities. The system can be used by environmental regulators and interest groups, one of which piloted field visits based on this system this past winter. Overall, our application demonstrates the potential for AI-based computer vision systems to solve major problems in environmental compliance with near-daily imagery. 
\end{abstract}

\begin{document}
	
\author{Ben Chugg}
\affiliation{
	\institution{Stanford University}
	\country{Stanford, CA, USA}
}
\authornote{Equal Contribution}
\email{benchugg@law.stanford.edu}

\author{Nicolas Rothbacher}
\affiliation{
	\institution{Stanford University}
	\country{Stanford, CA, USA}
}
\authornotemark[1]
\email{nsroth@law.stanford.edu}

\author{Alex Feng}
\affiliation{
	\institution{UC Berkeley}
	\country{Berkeley, CA, USA}
}
\email{alexfeng2000@berkeley.edu}

\author{Xiaoqi Long}
\affiliation{
	\institution{The California Institute of Technology}
	\country{Pasadena, CA, USA}
}
\email{xlong@caltech.edu}

\author{Daniel E. Ho}
\affiliation{
	\institution{Stanford University}
	\country{Stanford, CA, USA}
}
\email{deho@law.stanford.edu}

\maketitle

\section{Introduction}

How can advances in machine learning be used to transform environmental sustainability?  We illustrate its potential in agriculture, which is the leading contributor to water pollution in the United States~\cite{usepa2003national}. Of particular concern are Concentrated Animal Feeding Operations (CAFOs) (aka intensive livestock farms or ``factory farms''), industrial-scale livestock operations that raise large volumes of animals in close confinement. In 2008, CAFOs were responsible for more than 50\% of the total livestock production in the U.S. \cite{gurian2008cafos}, a number that has continued to grow in recent years \cite{hribar2010understanding}, both nationally and internationally~\cite{liu2018review}.

CAFOs produce substantial amounts of manure and manure-contaminated wastewater. The U.S.\ Government Accountability Office (GAO) notes that a large hog farm can produce more than 1.6M tons of manure annually, more than 1.5 times the amount produced by Philadelphia (1.5M residents)~\cite{mittal2009concentrated}. A large cattle farm can produce more than 2M tons of manure~\cite{mittal2009concentrated}. In 1998, CAFOs produced more than 133M tons of solid manure overall each year, more than 13 times the annual amount of human waste across the United States~\cite{burkholder2007impacts}. Unfortunately, more up-to-date estimates are difficult to obtain because, according to the GAO, ``[n]o federal agency collects accurate and consistent data on the number,
size, and location of CAFOs''~\cite{mittal2009concentrated}. However, the increase in the number of CAFOs tell us that the amount of waste is likely to have increased proportionally. 
Given the amount of waste produced by CAFOs, ensuring it is handled appropriately is of the utmost importance. 

CAFO waste can be a useful natural fertilizer. When applied injudiciously, however, it can lead to severe environmental and public health problems~\cite{bradford2008reuse}. Waste products have high concentrations of nitrogen and phosphorus, leakage of which into waterways causes eutrophication, i.e., deoxygenation and the resulting death of animal life from excessive vegetation growth~\cite{mallin2003industrialized}.  Even at recommended application rates, nearby streams can be contaminated by fecal matter~\cite{mallin2015industrial}. Other contaminants detected in both nearby groundwater and bodies of water include heavy metals (e.g., arsenic)~\cite{liu2015arsenic,nachman2005arsenic}, antibiotics~\cite{campagnolo2002antimicrobial}, microbial pathogens~\cite{gerba2005sources}, and endocrine disrupting hormones (e.g., steroidal estrogen)~\cite{raman2004estrogen,hanselman2003manure}. The effects are not limited to waterways. Application results in air emissions of ammonia, hydrogen sulfide, and methane~\cite{merkel2002raising}. The former are associated with health risks such as bronchitis, inflammation, and burns to the respiratory tract, while the latter increase greenhouse gas concentrations~\cite{hribar2010understanding}. 

\begin{figure}
    \centering
    \includegraphics[width=0.8\linewidth]{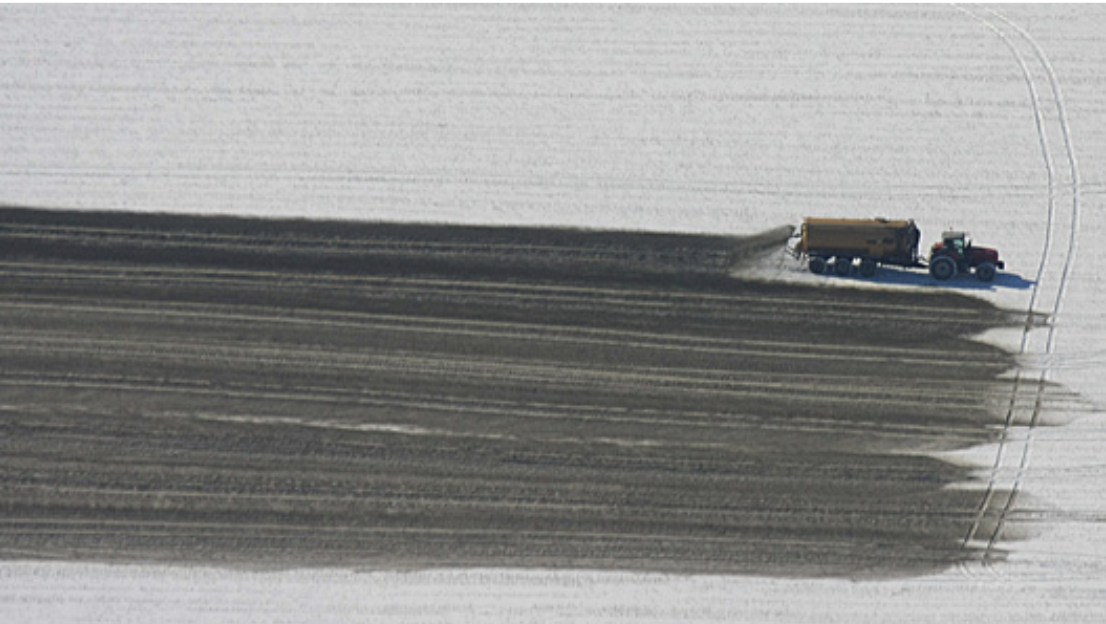}
    \vspace{-0.15in}
    \caption{Ground level view of truck performing land application in the winter. Image Credit: John Klein, Environmentally Concerned Citizens of South Central Michigan (ECCSCM). }
    \label{fig:truck}
\end{figure}

Of acute policy concern has been \emph{land application} (i.e., the distribution of manure on land) in the winter time~\cite{liu2018review, srinivasan2006manure}. Figure~\ref{fig:truck} illustrates the common technique for land application to a snow covered field, namely by use of a tractor-drawn tank spreader. Such winter land application poses distinct risks. Frozen and wet ground and the dearth of crops (which can take up manure nutrients) are believed to exacerbate nutrient runoff, thereby heightening the risk of ground and surface water pollution~\cite{klausner, lewis2009winter,srinivasan2006manure, williams2011manure}. The issue has been particularly contentious in Northern parts of the United States and Canada~\cite{srinivasan2006manure}. The International Joint Commission by the
Great Lakes Water Quality Board suggested that land application in the presence of snow and heavy rain be banned in order to keep algal blooms under control ~\cite{bihn2019oversight}. 

\subsection{Policy Setting}
\label{subsec:policy}

Due to the risks associated with winter land application, many efforts have been focused on how to regulate practices and monitor violations~\cite{liu2018review}. Transporting or storing waste is expensive and facilities are often limited in their storage capacity~\cite{bradford2008reuse}, such that some degree of winter application is likely. 

Most developed countries regulate or ban winter manure application~\cite{liu2018review}. The federal  Environmental Protection Agency (EPA) requires that permitted CAFOs follow a nutrient management plan (NMP) which determines where and how often they can apply. The specifics of the NMP are determined at a state level and there is significant heterogeneity in whether and how states regulate  winter land application~\cite{liu2018review}. In Wisconsin, where this project is focused, application of waste is restricted in the winter time.  In February and March, liquid waste may generally not be applied and solid waste may not be applied to any field that contains snow or is frozen~\S~NR 243.14(6)(c). Outside of February and March, liquid waste remains prohibited on frozen ground absent emergencies (\S~NR 243.14(7)(a)) and  application of solid manure remains restricted based on the amount of snow and slope of the field (\S\S~NR 243.14(6)(c) \& 243.14(7)(c)). 

Wisconsin law also provides for ``emergency application'' for liquid manure in cases of ``unusual weather conditions, equipment failure or other unforeseen circumstances beyond the control of the permittee.'' Such emergency land applications require prior verbal  approval from the Wisconsin Department of Natural Resources (DNR), followed by a written description of the event to the department within five days of its occurrence (\S\S~NR 243.14(7)(d)(1)(c)-(d)). 

All instances of land application are also required to be reported in detail. Facilities must file annual spreading reports, which  include dates of application, field information, and nutrient information. Such reporting  applies to ``surface applications on frozen or snow-covered ground,'' which must also report ``whether any applied manure or process wastewater ran off the application site'' (\S~NR 243.19(3)(c)(5)). 

While such self-reporting is required in principle, many have critiqued existing requirements as lacking teeth.  One article described EPA's then-rules as ``an exercise in unsupervised self-monitoring'' \cite{jerger2004epa}.  Verifying and ascertaining the extent of winter application -- throughout February and March specifically -- is difficult. Nor is it easy to understand the volume of approval for emergency applications. In response to a Freedom of Information Act request, the Wisconsin DNR indicated that it was not possible to release a complete list of authorized instances of land applications, as such information was not available in any database format. In the few disclosed instances of email authorizations, such land application events were not included in annual spreading reports. Moreover, while some reports include addenda that indicate winter land application, insufficient detail is provided to determine the volume, location, and dates of application. In short, very little is known about compliance with, and emergency exemptions from, prohibitions of winter land application. 

As a result, local environmental interest groups and motivated residents have taken the initiative to report suspected instances of application by physical visits and documentation (see, e.g., \cite{fieldsoffilth}). But the information asymmetry has made enforcement of the Clean Water Act -- and NMP permit terms specifically -- difficult. CAFO winter land application raises acute challenges for environmental regulation, as we do not understand basic questions: how often does winter land application occur? How many facilities engage in such practices?  How compliant are CAFO operators with state law and guidelines?  

\subsection{Our contribution}
To help determine the extent of winter application, we build a real-time detection system which pulls daily 3m/pixel satellite imagery at known CAFO locations and detects land application using convolutional neural networks (CNNs). 
We analyze our approach on 330 CAFOs across the state of Wisconsin, where land application in violation of permit terms has long been suspected.  
We also use our method to gain insights into how and when application occurs, as well as variation between facilities. For instance, we find facilities which increase the number of winter application by over 150\% in Winter 2020 compared to 2018 and 2019. We also discover that some facilities have some application on the ground nearly half of the winter season, and that a significant number of application events (20\%), occur within one day of each other.  Our model also determines that a substantial number of application events occur within a short time frame of one another. 

In sum, our contributions are fivefold. First, we introduce the highly consequential environmental, policy, and agricultural setting of CAFOs and land application. Second, we provide a new dataset consisting of time series of satellite imagery at 330 verified CAFO locations across Wisconsin from 2018-2020 (n=57,697), with 1,813 instances of application identified across 96 of these facilities. Third, we develop an automated system to download satellite imagery on a daily or weekly basis and conduct inference on each image for likelihood of land application. Fourth, we prototype a method to convert image-by-image predictions into predictions of application \emph{events} (which may span more than one image). Fifth, we conduct a retrospective estimate of the prevalence of land application events when presumptively prohibited (February - March 2022) and show that the rate of land application remains alarmingly high during these months. 

Our land application detection system was used in partnership with the Environmental Law and Policy Center (ELPC) at the end of the 2021/22 winter Season to investigate several possible instances of application. The partnership is set to continue throughout the 2022/23 winter season, when the system will be used to dispatch volunteers on a weekly basis to identify possible land application.

\begin{figure}
\centering 
\begin{minipage}{0.49\textwidth}
\subcaptionbox{Location A, Before}{\includegraphics[scale=0.23]{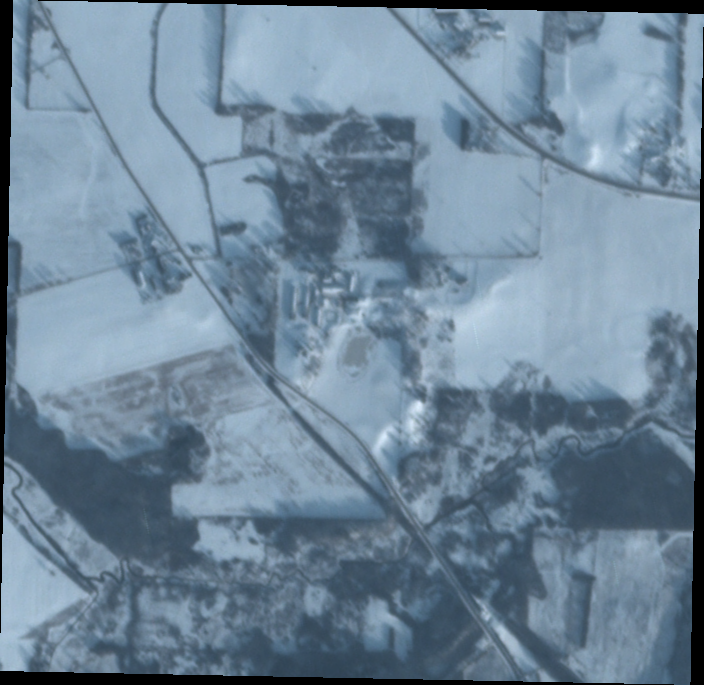}}
\subcaptionbox{Location A, After}{\includegraphics[scale=0.23]{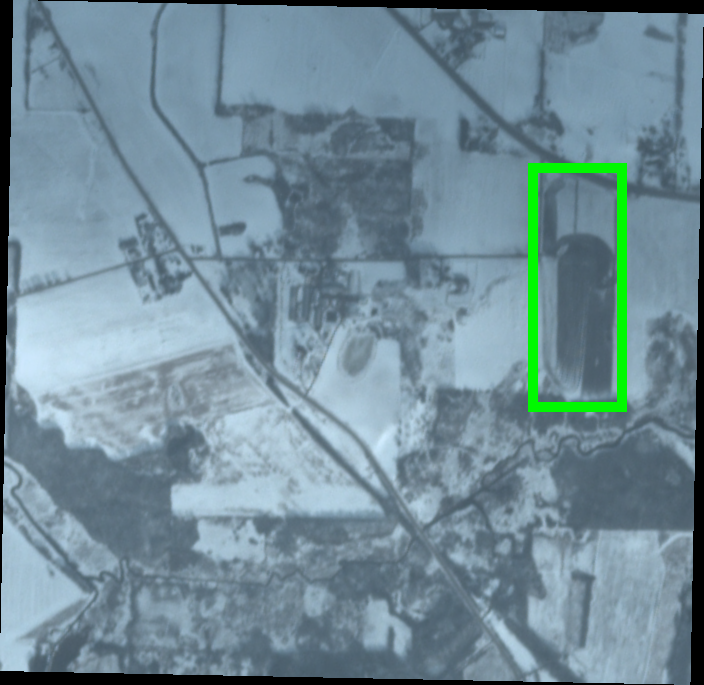}}
\end{minipage}
\begin{minipage}{0.49\textwidth}
\subcaptionbox{Location B, Before}{\includegraphics[scale=0.23]{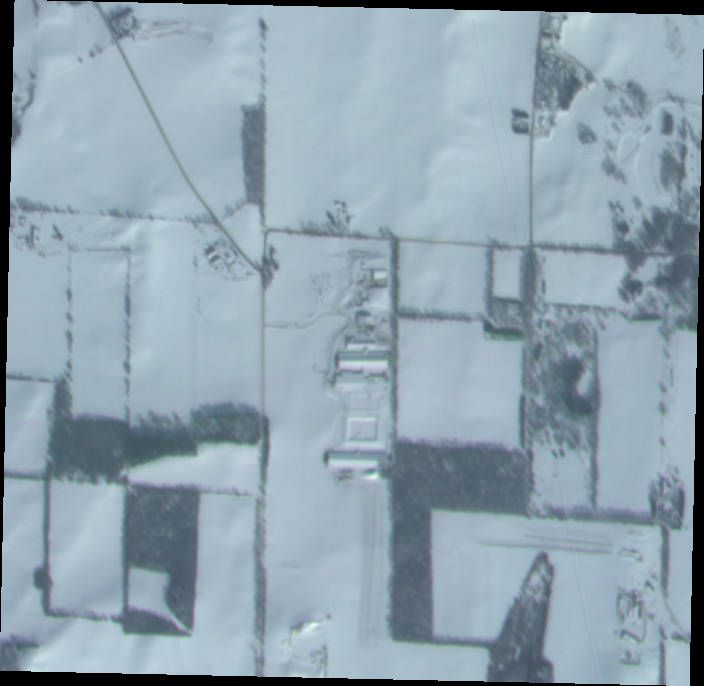}}
\subcaptionbox{Location B, After}{\includegraphics[scale=0.23]{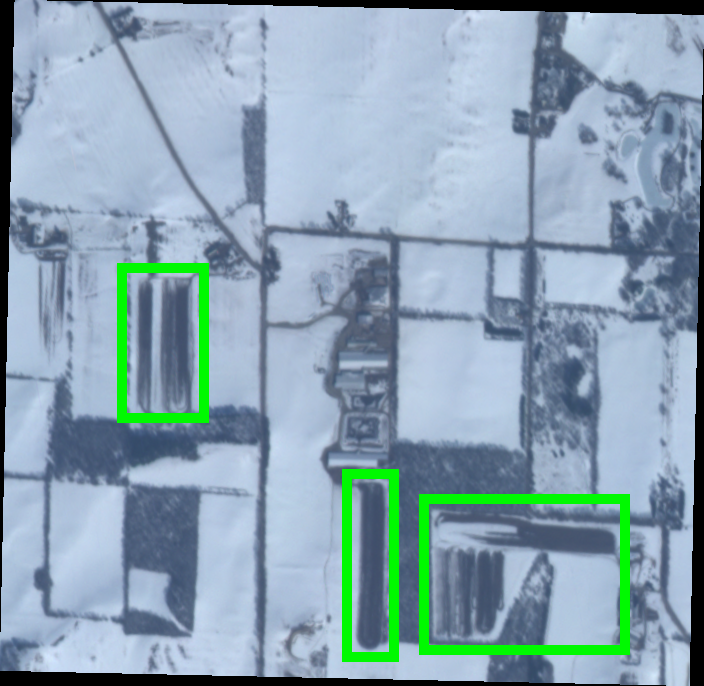}}
\end{minipage}
\vspace{-0.15in}
\caption{Examples of land application in 3m/pixel resolution satellite imagery.  Images (b) and (d) show examples of application (enclosed by green rectangles) that was not present in images (a) and (c).  }
\label{fig:app_examples}
\end{figure}

\subsection{Related Work}

Our work relates to four bodies of research. First, the environmental risks posed by CAFOs have given rise to recent work on automating their detection and analysis. Computer vision techniques have been used to classify CAFOs by type in North Carolina \citep{handan2019deep}, and to detect poultry CAFOs nationwide \citep{robinson2021mapping,maroney2020using,patyk2020modelling}. Approaches have also been developed to detect when a given CAFO shed was built \citep{montefiore2022reconstructing,robinson2021temporal} and whether a facility has expanded over time \citep{chugg2021enhancing}. This prior work is an important starting point for using computer vision in environmental sustainability. Prior work, however, has not utilized near real-time satellite imagery to detect likely violations of permit terms, as we do here. 

Second, our work relates more broadly to the literature on satellite detection for the environment. Examples include forest fire detection~\cite{yu2005real}, mapping crop cycles~\cite{gao2021mapping}, and detecting droughts~\cite{verbesselt2012near}. In general, there is an increasing awareness as to how modern computer vision techniques, and deep learning in particular, can be applied to satellite imagery to tackle environmental and regulatory challenges~\cite{handan2021deep}. Our paper also bears some relationship to previous work in anomaly detection in aerial imagery, such as detecting failures in power plants~\cite{vlaminck2022region} and building damage detection~\cite{tilon2020post}. Indeed, it is straightforward to cast our problem in terms of anomaly detection. Application events themselves may considered anomalies, as they appear suddenly from one image to the next. But behavior at the facility level may also be anomalous. For example, there may be a drastic increase in the number of application events from one year to the next. See Section~\ref{sec:anomalies} for results on this front. 

Third, on the technical side, our work builds off the rapid evolution of and progress in deep learning for computer vision. Both objection detection and image classification have been increasingly utilized with satellite and aerial imagery, see \citet{cheng2016survey} and \citet{dhingra2019review} for excellent surveys. Our method is based off of the You-Only-Look-Once (YOLO) family of object detection models~\cite{redmon2016you,bochkovskiy2020yolov4,redmon2018yolov3}. 

Last, our introduction of this setting of high environmental consequence contributes to the machine learning community. Many scholars have critiqued existing machine learning work as being too fixated on a narrow set of technical benchmarks, without sufficient engagement with real-world problems~\cite{church_kordoni_2022,Koch, wagstaff2012machine}. There is growing recognition that while pursuing state-of-the-art on well-defined and rigorously curated benchmark tasks can create better models, such progress does not necessarily translate to real world impact~\cite{raji2021ai}. Part of the reason lies in the significant barriers to collecting, ground truthing, and releasing meaningful real-world machine learning problems~\cite{handan2021deep, sambasivan}. We release this CAFO land application dataset and all accompanying code \footnote{\url{https://github.com/reglab/land-application-detection}}. Our contribution here illustrates the possibility of substantial impact of machine learning in environmental sustainability and governance and we hope to inspire the community to help solve these challenging problems~\cite{engstrom2020government,hino2018machine}. 

\section{Data}
\label{sec:data}

\begin{figure}
    \centering
    \includegraphics[scale=0.35]{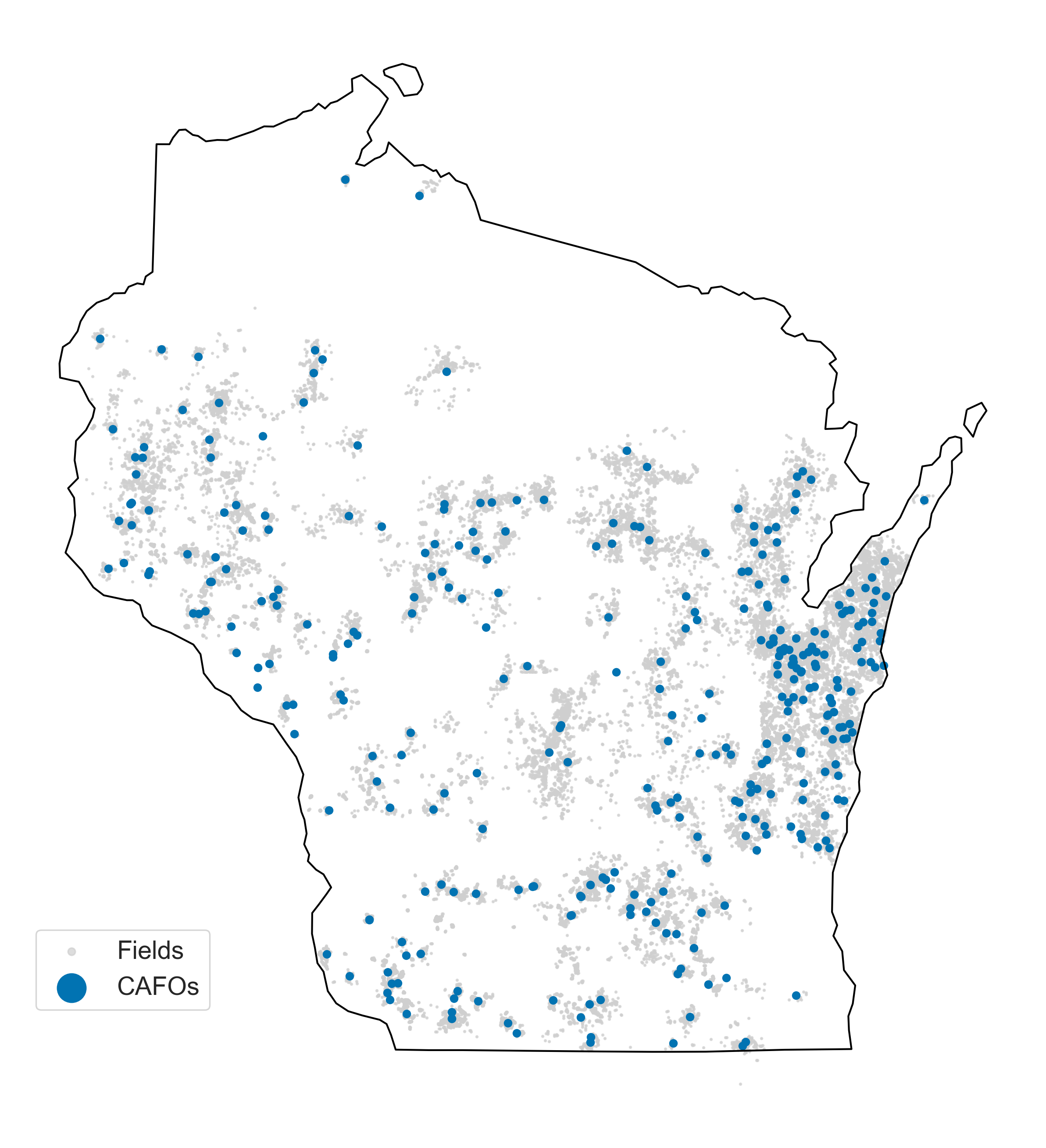}
    \vspace{-0.3in}
    \caption{Map of permitted fields for land application and permitted CAFOs in Wisconsin. Blue dots represent CAFO locations and grey areas indicate fields where land application is allowed subject to permit requirements. This map shows that land application areas are proximate to the CAFO facility.}
    \label{fig:map}
\end{figure}

We now describe how we created a new ground truth dataset on land application. 
First, we obtained the location of 330 CAFOs in Wisconsin from the Environmental Working Group (EWG), which spent many months manually scanning satellite imagery to identify CAFO locations. Through a Freedom of Information Act request with WI DNR, we also obtained shape files for permitted land application areas. Figure \ref{fig:map} plots the CAFO locations and permitted land application areas. Because transportation of waste is expensive, most land application occurs proximate to the CAFO facility. 

Second, we acquired daily satellite imagery at 3m/pixel resolution through a subscription with  PlanetLabs~\cite{planet}. For each CAFO location, we pulled imagery from the 2018, 2019, and 2020 winter seasons (i.e., from November 1 of the given year to March 1 of the subsequent year). The principal rationale is that land application both poses greater environmental harm in the wintertime and is restricted, particularly in February - March, under Wisconsin law (see Subsection~\ref{subsec:policy}). In addition, detecting land application from 3m/pixel resolution is a hard task, but it is much easier to train humans to determine such instances under snow cover and dates outside that range rarely contained snow. We downloaded images that were 1km by 1km squares, centered on the CAFO locations. 

Figure~\ref{fig:app_examples} provides examples of the Planet Imagery. The left column presents CAFO locations and surroundings before land application. The right column presents those same facilities with land application events, which resemble the kind of tractor-drawn spreading events depicted in Figure~\ref{fig:truck}. We filtered the images by cloud cover, clarity, existence of image artifacts, and snow cover. This process left us with approximately one image per week per location. 

Third, based on a sample of images provided by subject matter experts, we trained a team of undergraduate students and staff researchers to label imagery for instances of land application. This training involved extensive consultation and learning about agricultural practices over the course of a summer, drawing on experts at the Socially Responsible Agriculture Project and the Environmental Law and Policy Center (ELPC). Land application areas within images were labeled with a bounding box, resulting in a positive label for every instance where an application event is captured within a Planet satellite image. To ensure label quality, each image was labeled by at least two members of the research team.  This resulted in the identification of 1,061 images containing application and 1,813 total instances of application across 96 locations.\footnote{44\% of positive images had multiple instances of land application.} Although there are strong reasons to believe that reports are incomplete -- particularly with respect to winter -- we also examined a small sample of annual spreading reports and confirmed that over 40\% of the reported events were visible in our data.\footnote{A major reason why other reported instances may not be available stems from the filtering for high quality satellite imagery.}

Fourth, to construct application \emph{events} from the image-level bounding boxes, we consider a single event to be a series of application boxes which have non-zero intersection across an unbroken series of images ordered by time. This leaves us with 151 distinct land application events.

\section{Problem Formulation}
\label{sec:formulation}

Given a series of images of the same location over time (described in Section~\ref{sec:data}), our goal is to (i) predict whether each image contains an instance of land  application, and (ii) identify the unique application \emph{events} in the series. Note that an event may (and probably does) persist across images, since application does not fade immediately. 

Formally, we have a set of locations $L$ and years $Y$.  For each location-year pair $(\ell,y)\in L\times Y$, we have a series of 3-band images $X_1,\dots,X_N$ and events $E_1,\dots,E_K$. An event $E_k$ is a tuple $(k_1,k_2,B_k)$ where $k_1\in\{1,\dots,N\}$ is the index of the image in which the event starts, $k_2$ the index at which it ends, and $B_k$ is a bounding box enclosing the application event in the images (it is the same for each image since each image has the same coordinates). Of course, a location-year pair might have no events.

As noted before, we consider $E_k=(k_1,k_2,B_k)$ to be an event of images $X_{k_1},X_{k_1+1},\dots,X_{k_2}$ containing overlapping labeled application boxes, and $X_{k_1-1}, X_{k_2+1}$ do not have such boxes (if they exist). Then we set $B_k$ to be the union of all the application boxes across $X_{k_1},\dots,X_{k_2}$.  
Our three tasks are: 

\begin{enumerate}
    \item[{\bf T1.}] (Task 1 -- Image classification) For each image $X_i$, determine whether it contains one or more instances of land application (image classification).
    \item[{\bf T2.}] (Task 2 -- Object Detection) For each image $X_i$, draw a bounding box $B$ around any application. 
    \item[{\bf T3.}] (Task 3 -- Event Detection) Detect the application events $E_1,\dots,E_K$. 
\end{enumerate}

Task 1 is useful for real-time detection of application, and prioritization of enforcement resources on a daily or weekly basis. Task 2 is useful for identifying precisely where in the image the application lies, and for aggregating image-level predictions into event detections (Task 3).  
Task 3 is useful for gathering descriptive statistics on past data. For instance, to estimate the total amount of application, we typically care about the number of distinct application events, not merely the number of images containing application. The latter can be influenced by exogenous factors such as a fresh snow fall or cloud cover and is thus not a good proxy for the total amount of application. Moreover, as demonstrated by Figure~\ref{fig:app_examples}, there can be multiple instances of  application in a single image. Thus, classification can undercount the true number of application events.

\section{Methods}
\label{sec:methods}

We consider four distinct models: Two image classification methods (for Task 1), and two object detection methods (for all three tasks), all of which use CNNs. Recall the distinction: image classification methods predict a label for the image as a whole (in our case a binary label, application/no-application), while object detection predicts bounding boxes enclosing predicted classes within each image. Even though the image classification methods cannot be applied to Tasks 2 or 3, we consider them useful baselines. In principle, environmental regulators and interest groups may want to have separate models for different tasks since CNNs designed specifically for image classification may perform better on Task 1 than object detection methods. 

\subsection{Evaluation}
The image classification methods are built for Task 1. However, we also evaluate the object detection models on Task 1. To do so, we need to extract image level classification results from bounding boxes. We do this in the natural way: For any image in which the object detection methods predicted a bounding box, we set the predicted label to True. We set the model score for that image to be the maximum confidence over all detections in the image. If the image contained no detections, we set the predicted label to False, and the model score to be 0. In this way we can evaluate the precision and recall for Task 1. 

To evaluate the object detection methods on Task 3 (event detection), we aggregate image-level bounding box predictions into event predictions by considering an event to be any series of detections in an unbroken sequence of images whose bounding boxes all overlap. We compare these predictions to the true predictions as follows. For each true event $E_k$ we ask whether there is a predicted event $E_j'$, any of whose bounding boxes intersect with $B_k$ and whose dates intersect those of $E_k$. That is, $E_j'$ cannot have begun after $E_k$ ended or vice versa.

We train the four models with a 70\%/10\%/20\% train/validation/test split. We split the data by location because the dataset contains multiple images per location per year. A random split would thus likely lead to images of the same location in both the train and test sets, potentially overfitting and/or inflating model performance based on idiosyncrasies of specific locations.

\subsection{Image Classification Models}
We consider two baselines for Task 1, both employing pretrained CNNs trained to predict whether there exists application in an image, i.e., as a binary classifier for each image. We use an Xception \cite{chollet2017xception} CNN backbone pretrained on ImageNet and fine-tuned on our data. During training, we allow the layers of the pretrained network to be optimized. Optimization was performed with the SGD optimizer with learning rate 0.0005. We activate all nodes using the ReLU function and use a dropout rate of 0.5 for all layers. The best model was selected based on performance on a validation set of 10\% of the labeled locations. Both baseline models were implemented in Keras \cite{chollet2015keras}.

\paragraph{\bf Single CNN} The first baseline is a single Xception model with a single convolutional layer added before the pretrained layers. The outputs are run through max pooling to match the size of the input layer of Xception. The outputs from the Xception convolutional layers are flattened to one dimension using global average pooling. We remove the fully connected layers of Xception and replace them with a single fully connected hidden layer and an output layer with one node. The logistic scores from that layer are passed through a sigmoid activation to make the predicted confidence of the input image containing application.

For the single CNN model, the training data was down sampled so that positives and negatives made up equal proportions of the training set. The best model was trained for 146 epochs.

\begin{table*}[t]
    \centering
    \begin{tabular}{r|ccccc|cc}
    & \multicolumn{5}{c|}{\bf Image Classification} & \multicolumn{2}{c}{\bf Event and Object Detection} \\ 
    {\bf Model} & PR AUC & ROC AUC & $F_1$@0.5  & $F_2$@0.5 & $F_{0.5}$@0.5 & PR AUC & mAP@0.5 \\
    \hline 
    \emph{Single CNN} &  0.84 & 0.87 & 0.65& 0.56 &0.78& - & -\\
    \emph{Dual CNN}&  0.71 & 0.74 & 0.12 & 0.09 & 0.27& - & -\\
    \emph{Faster R-CNN} & 0.92 & 0.92 & {\bf 0.77} & {\bf 0.88} & 0.67& {\bf 0.63} &  {\bf 0.57} \\
    \emph{YOLOv5}&  {\bf 0.94} & {\bf 0.93} &  0.76 &0.68 & {\bf 0.87} & {\bf 0.63} &  {\bf 0.57}
    \end{tabular}
    \caption{Results for each method on both Tasks 1, 2 and 3 (where applicable). For each metric, we bold the results with the best performance. The``@0.5'' next to the $F_\beta$ scores indicates that the classification threshold is at 0.5. PR AUC is the area under the precision-recall curve, and ROC AUC is the area under the receiver operating characteristic curve. The (mean) average precision for object detection is performed at an IoU threshold of 0.5.  }
    \label{tab:model_accuracy}
\end{table*}

\paragraph{\bf Dual CNN} The second baseline compares \emph{pairs} of images, attempting to detect whether application is present in the latter. This baseline is motivated by the observation that, for humans labelers, time-series information was often helpful. Separating application from other features (e.g., trees, dark sheds, roads) is easier if one can see whether purported application is a consistent feature across time. The dual CNN approach is inspired by several successful applications of this methodology to a wide range of tasks, such as fracture detection~\cite{choi2020using}, optometry~\cite{repala2019dual}, and analyzing EEG~\cite{cai2020graph}.

The dual input model uses a single Xception backbone that is applied to both images at the same time. The model uses a similar input architecture to the single CNN (a CNN layer with max pooling). The two Xception global average pooled outputs are concatenated into one flat vector which is passed into a fully connected network with one hidden layer and a single output node. The logistic scores from that output node are passed through a sigmoid function to make the predicted confidence of the second of the input images containing application. 

For the dual input model, the training samples labeled as positives were upsampled by three times and the negatives were downsampled so that the negatives and positives would compose equal proportions of the training set. The resulting best model was trained for 48 epochs.

\subsection{Object Detection Models}
For both object detection models we perform lightweight hyperparameter optimization via a grid search across the learning rates. We try rates of 1e-4, 2.5e-4, 5e-4, 0.001, 0.005, and 0.01. Beyond that, we engage in limited hyperparameter tuning in an attempt to demonstrate that environmental interest groups and regulators can employ off-the-shelf methods to great effect. 

\paragraph{\bf YOLOv5}
Our object detection model is the  You Only Look Once (YOLO) \citep{redmon2016you} network, trained on ImageNet and fine-tuned on our single class application dataset. We use Ultralytics implementation of YOLOv5. We use a learning rate of 0.01, momentum 0.937, and weight decay of 0.0005. We ensure the anchor boxes are well-fit to the data before training. The network was trained for 100 epochs.

\paragraph{\bf Faster R-CNN} For Task 2, we use faster R-CNN as a baseline~\cite{ren2015faster}. We use the Detectron2 implementation~\cite{wu2019detectron2}, pretrained on ImageNet and fine tuned on our single class application dataset. The convolutional backbone we use is ResNet-50 \cite{He2015resnet} with a fully connected faster R-CNN head for object detection. During fine-tuning, we use momentum 0.9, weight decay 0.0001, learning rate 0.000025 with a linear learning rate warm up over the first 1000 iterations. The best model was trained for 10,000 iterations.

\section{Results}

\subsection{Accuracy}
Table~\ref{tab:model_accuracy} summarizes the performance statistics across models and tasks. Regarding image classification (Task 1), we find that both object detection methods (YOLOv5 and Faster R-CNN) outperform both of the networks designed specifically with classification in mind. The areas under the precision-recall (PR) and receiver operating characteristic (ROC) curves 
for YOLOv5 and Faster R-CNN are 0.94, 0.93 and 0.92, 0.92 respectively. This is compared to 0.84 and 0.87 for the Single CNN model, and 0.71 and 0.74 for the dual CNN. Somewhat surprisingly, the dual CNN fairs worse than the single CNN, implying the extra temporal information did not help. 

We also report the $F_\beta$ scores for $\beta\in\{0.5,1,2\}$ at a classification threshold of 0.5. We note that the parameter $\beta$ determines how much weight is placed on recall relative to precision. We report all three since environmental interest groups and regulators may weigh objectives differently. For instance, some might prefer to detect as much application as possible (weighting recall higher than precision), while others may want to ensure that field visits or inspections are  successful (weighting precision higher than recall). While Faster R-CNN and YOLOv5 have similar $F_1$ scores (0.77 and 0.76 respectively), they differ in the others. Faster R-CNN has an $F_2$ score of 0.88 and $F_{0.5}$ score of 0.67, while the performance of YOLOv5 is nearly reversed: 0.68 $F_2$ score and $0.87$ $F_{0.5}$. YOLOv5 thus trades off recall for higher precision relative to Faster R-CNN. 

For Task 2 (object detection), the mean average precision (mAP) (equivalent, average precision, since there is a single class) for both object detection methods at an Intersection over Union (IoU) threshold of 0.5 is the same at 0.57. 
For Task 3 (event detection) both YOLOv5 and Faster R-CNN achieve a AUC of 0.63 on the PR Curve, demonstrating that this task is indeed more difficult than Task 1. We do not give ROC statistics for this task because the number of true negatives is not well-defined.  

\begin{figure}[tb]
    \centering
    \begin{minipage}{0.21\textwidth}
    \subcaptionbox{TP (Score 0.84)}{\includegraphics[width=1.2in,height=1.2in]{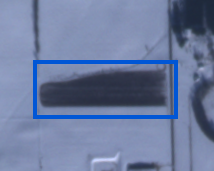}}
    \end{minipage}
    \begin{minipage}{0.21\textwidth}
    \subcaptionbox{TP (Score 0.71)}{\includegraphics[width=1.2in,height=1.2in]{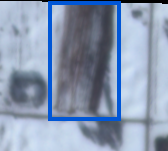}}
    \end{minipage}
    \begin{minipage}{0.21\textwidth}
    \subcaptionbox{TP (Score 0.93)}{\includegraphics[width=1.2in,height=1.2in]{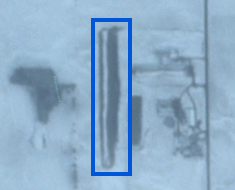}}
    \end{minipage}
    \begin{minipage}{0.21\textwidth}
    \subcaptionbox{FP}{\includegraphics[width=1.2in,height=1.2in]{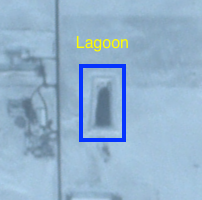}}
    \end{minipage}
    \begin{minipage}{0.21\textwidth}
    \subcaptionbox{FP}{\includegraphics[width=1.2in,height=1.2in]{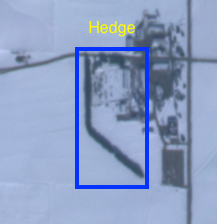}}
    \end{minipage}
    \begin{minipage}{0.21\textwidth}
    \subcaptionbox{FN}{\includegraphics[width=1.2in,height=1.2in]{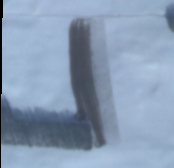}}
    \end{minipage}
    \vspace{-0.1in}
    \caption{Examples of true and false model predictions (blue boxes). TP = True Positive, FP = False Positive, FN = False Negative. The images have been zoomed in on the area of interest. Images (a), (b), and (c) represent true positives at the standard classification threshold of 0.5. Images (d) and (e) are false positives. (d) correctly identifies application, but it is on a lagoon and is therefore not land application. (e) identifies a hedge partly enclosing the property. (f) provides an example of a false negative, as the vertical streaks are land application but are not identified as such by the model. }
    \label{fig:model_detections}
\end{figure}

Overall, we find that YOLOv5 and Faster R-CNN perform very similarly, and one can be justified in using either for both Tasks 1 and 2. We use YOLOv5 for the results in the remainder of the paper. Figure~\ref{fig:model_detections} gives several example of YOLO's predictions -- three true positives, two false positives, and one false negative (at varying classification thresholds).

\subsection{Trends and Outliers}
\label{sec:anomalies} 
While the previous section lends insight as to how the model performs as a real-time detection mechanism, we are also interested if it can be used to gather insights into application trends over the past few years. Consequently, we run our model over 130 CAFO locations for the 2018, 2019, and 2020 winter seasons that were not in our training or testing sets when developing the model. We restrict our analysis in this section to these unlabeled images to ensure findings are not driven by knowledge of the training data.  

\begin{figure*}[t]
    \centering
    \begin{minipage}{0.32\textwidth}
    \includegraphics[scale=0.4]{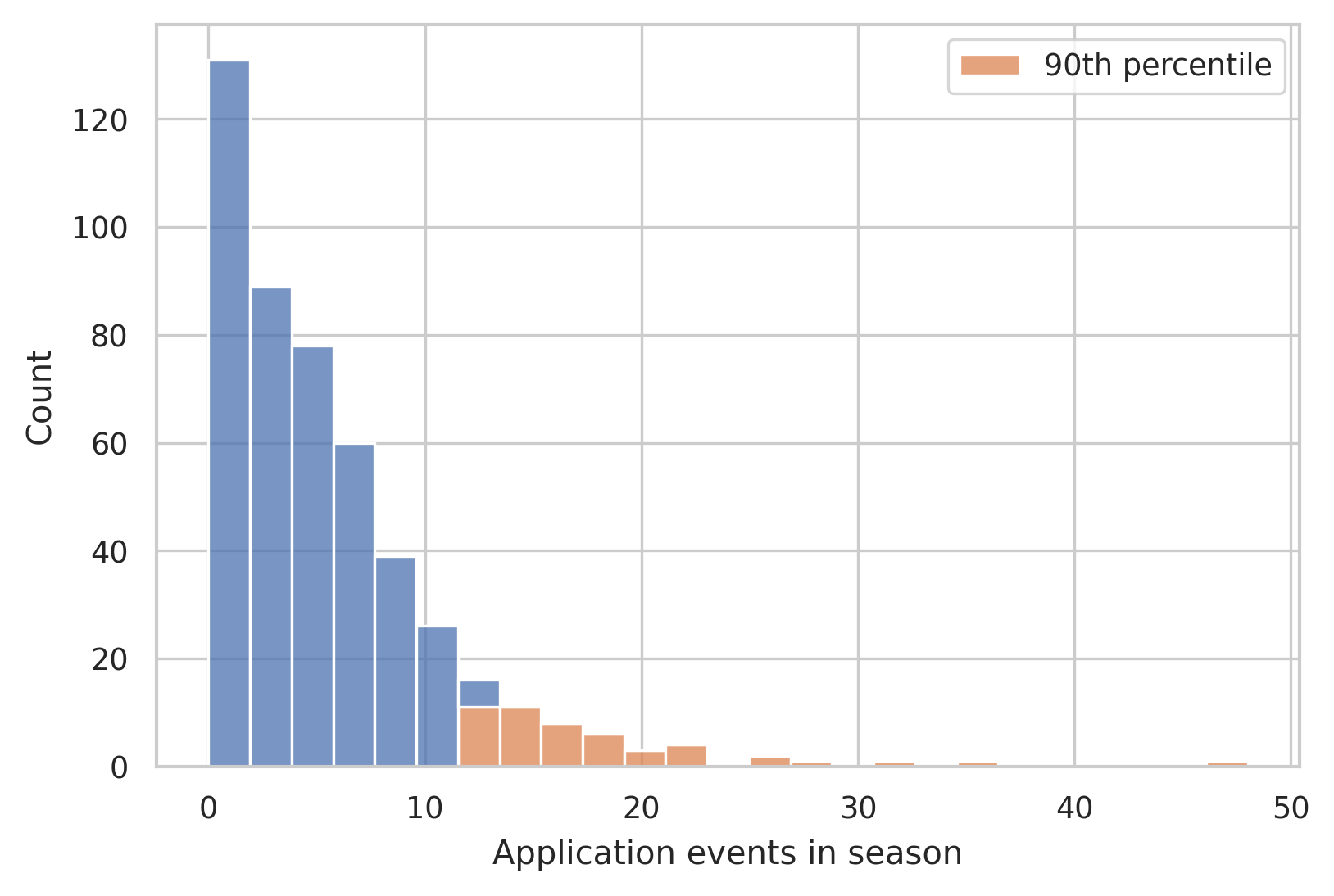}
    \end{minipage}
    \begin{minipage}{0.32\textwidth}
    \includegraphics[scale=0.4]{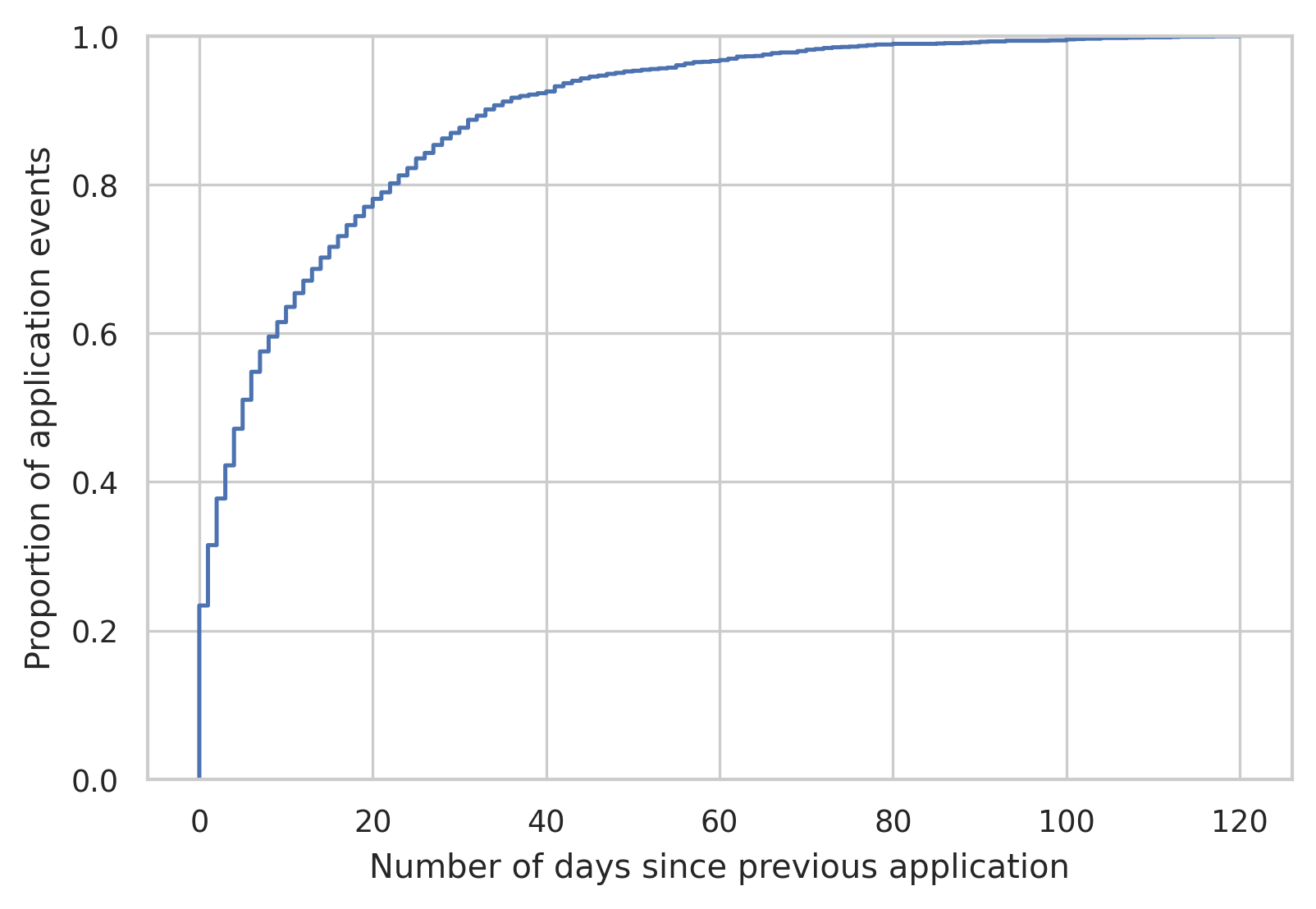}
    \end{minipage}
    \begin{minipage}{0.32\textwidth}
    \includegraphics[scale=0.4]{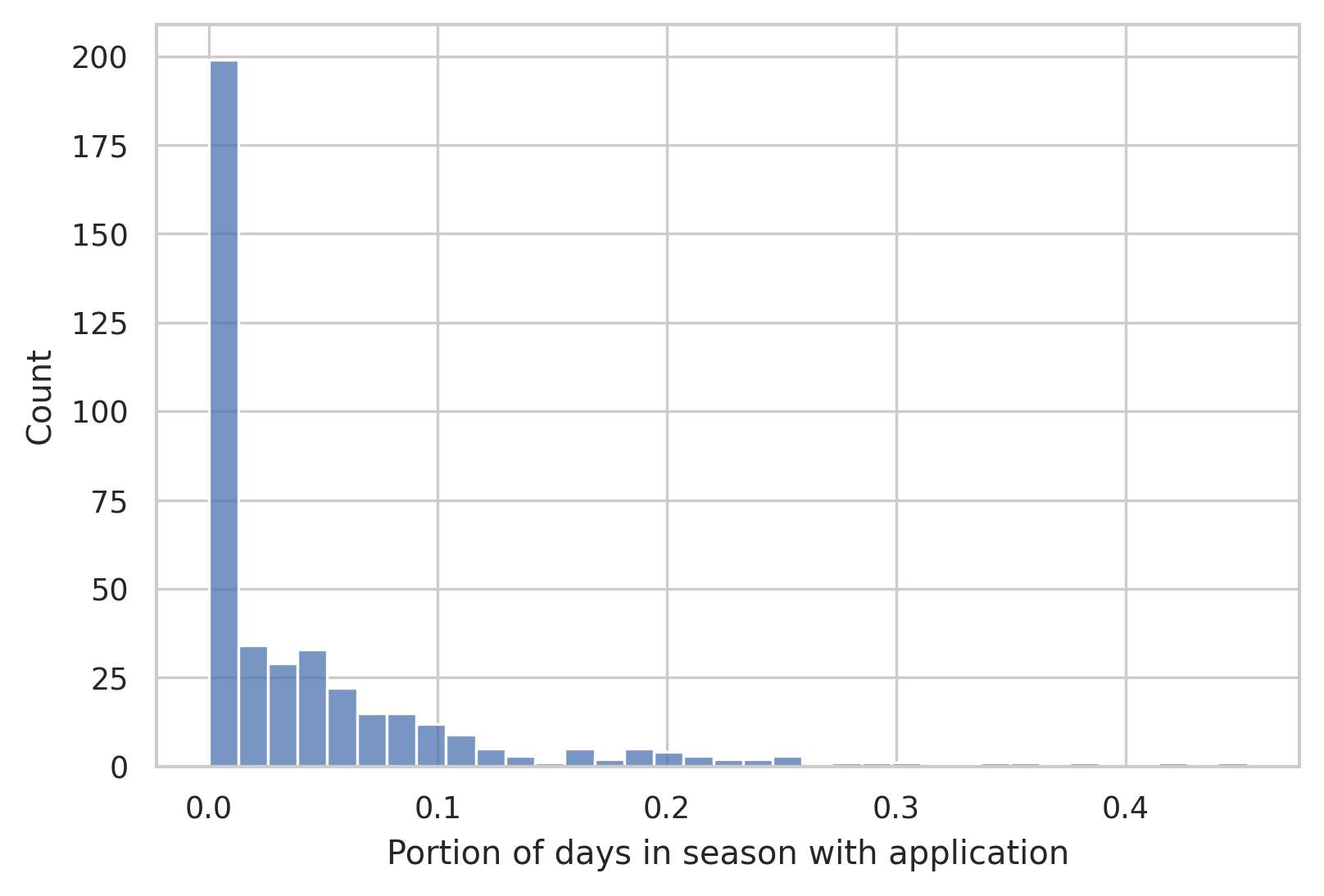}
    \end{minipage}
    \vspace{-0.1in}
    \caption{Left: Distribution of number of application events in a single season and location. Orange bars indicate the location-seasons in the 90th percentile. Center: Empirical cumulative distribution function of the time between subsequent application events in one location during. Right: Histogram of the percent of a location's season that was found to have visible application. }
    \label{fig:trends}
\end{figure*}

\paragraph{\bf Number of events per facility} Figure \ref{fig:trends} (left) plots the number of events detected per facility per season. The power law relationship suggests that some facilities are applying at much higher rates than the average. This allows us to identify outliers, which likely pose greater environmental risk: Looking at a facility in the right tail, we find a facility which applied 15 times in Winter 2018, 42 times in Winter 2019, and 16 times in Winter 2020,
when the average number of events across all facilities is 5.\footnote{We note that while repeated land application is likely  a high risk factor, it is also possible that such practices reflect attempts to apply in lower risk conditions (e.g., after snow has melted, in areas distant from waterways).} As a policy matter, the results suggest that the permit terms for such a facility may not support to intensity of livestock production, requiring larger manure storage systems, reduction in animal count, or larger land application area. Identifying such outliers using this approach can help to  allocate scarce compliance resources where needed most.

\paragraph{\bf Timing of application events} Figure \ref{fig:trends} (center) demonstrates how many days after one application event the next event occurs. Interestingly, 30\% of applications occur within the first day following a previous application event. In such instances, multiple instances of land application occurred in close proximity, as can be seen in Figure~\ref{fig:app_examples}.  55\% of subsequent events occur within the first week of the first event. If we examine the percentage of the season that will have an application event visible, it is clear that the vast majority of locations spend much of the season without applying. In some  locations, however, visible application exists in surrounding fields for 20-45\% of the Winter months.

\paragraph{\bf Number of events per month} In 2008, Wisconsin introduced the sharp prohibition on winter land application for February-March. We hence assess whether this sharper restriction in fact reduces application rates. Figure~\ref{fig:prevalence_and_time} (right) plots the time series of land application events, showing that the time series exhibits no reduction in application rates for those months. This suggests that the 2008 restriction, absent stronger enforcement measures, did little to reduce winter land application.

\subsection{Prevalence Estimation, February-March 2022}
\label{sec:prevalence}

We now use our model to estimate the total number of application events at the 330 CAFO locations in our dataset in February and March 2022. 
Our dataset restricts us to detecting application within a distance of 500m of each CAFO, and thus our prevalence estimates should be taken as lower bounds on the total amount of application.  

To obtain the prevalence estimate we run our model over each location throughout February and March and aggregate the image-level predictions into event predictions as per Section~\ref{sec:methods}. Instead of selecting an arbitrary confidence threshold and tallying the resulting event detections, we take advantage of the gradient of confidence scores to obtain a more reliable estimate. Additionally, this supplies us with more fine-grained insight into the calibration of our model.  

We stratify detections according to their confidence in bins $B_1,\dots,B_J$, with the detections in $B_i$ having lower confidence than those in $B_{i+1}$. Additionally, we let $B_0$ contain those images with no detections. If we sample $n_i$ items from bin $B_i$ and find $a_i$ true instances of application among those examples, then the estimate for the total prevalence is 
\begin{equation}
\label{eq:estimate}
    T = \sum_{i=0}^{J} \frac{a_i}{n_i}|B_i|,
\end{equation}
i.e., the sum of empirical probabilities multiplied by the size of each bin. The stratification of images by confidence further enables us to examine how the model's confidences compare to the obtained empirical probabilities. 
    
We run the model with a prediction threshold of 0.25, removing all detections with confidence score below this value. We cluster the image level predictions into events and then cluster into four bins: $B_0$ indicating images with no predictions; and $B_1$, $B_2$, and $B_3$ indicating predicted events with confidence in $[0.25, 0.5)$, $[0.5, 0.75)$, and $[0.75, 1]$, respectively.

We sample $B_0$ 120 times and $B_1,B_2,B_3$ 60 times each.
For each detected event, we check whether it was a true positive (i.e., a true event). To qualify as a true positive, it must (i) overlap with a true event, and (b) be the first detected event to do so. For example, if a true event is 10 days long, the first four days of which are detected as a single event by the model, and the last four as another, the second detected event is not counted as a positive. We also check how many images of the detected event contained application (e.g., image classification accuracy). In the previous example, while the second detected event is not counted as a novel event, all four images contain application.  

\begin{figure*}[t]
    \centering
    \begin{minipage}{0.49\textwidth}
    \centering
    \includegraphics[scale=0.4]{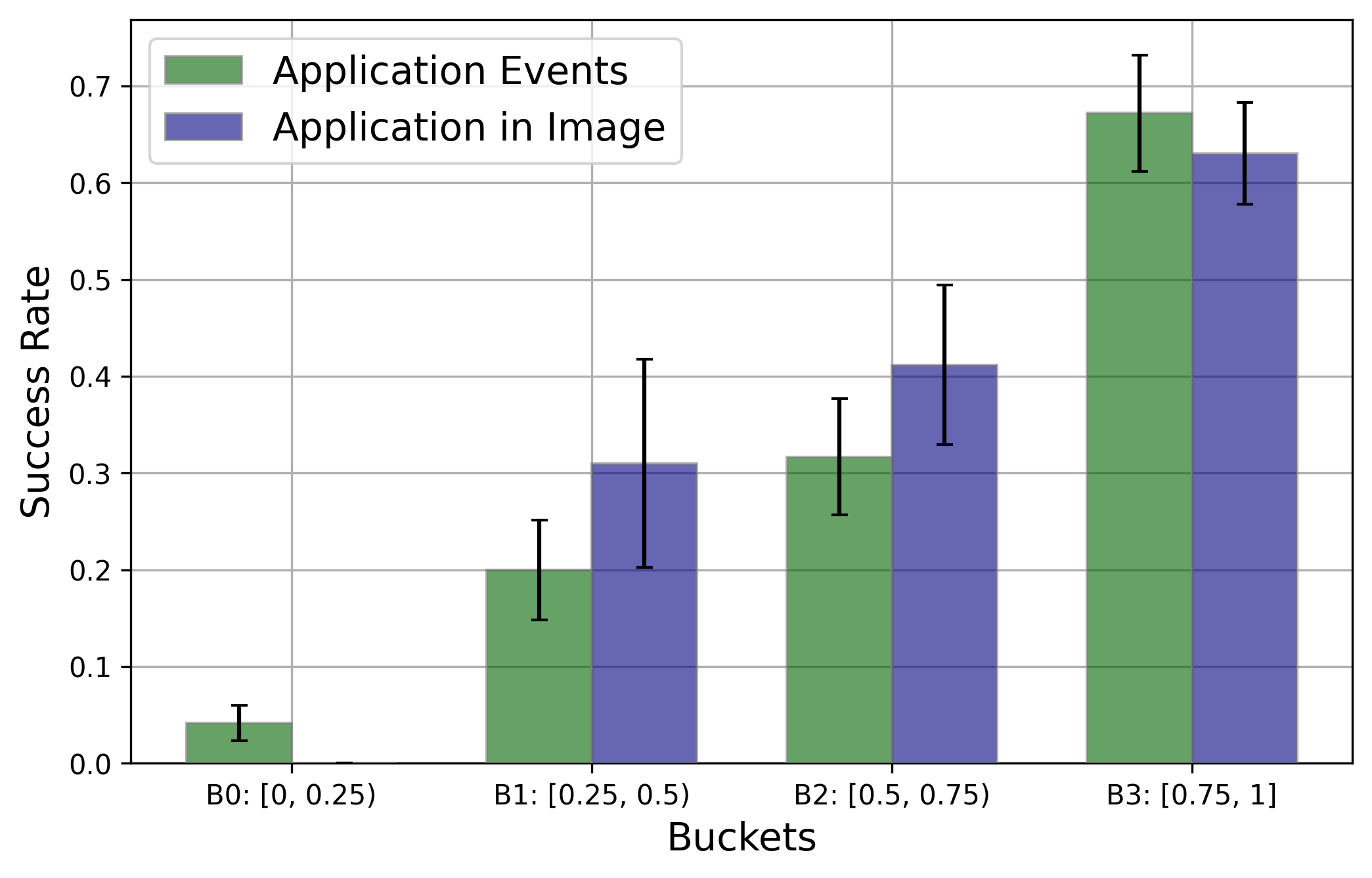}
    \end{minipage}
    \begin{minipage}{0.49\textwidth}
    \centering
    \includegraphics[scale=0.5]{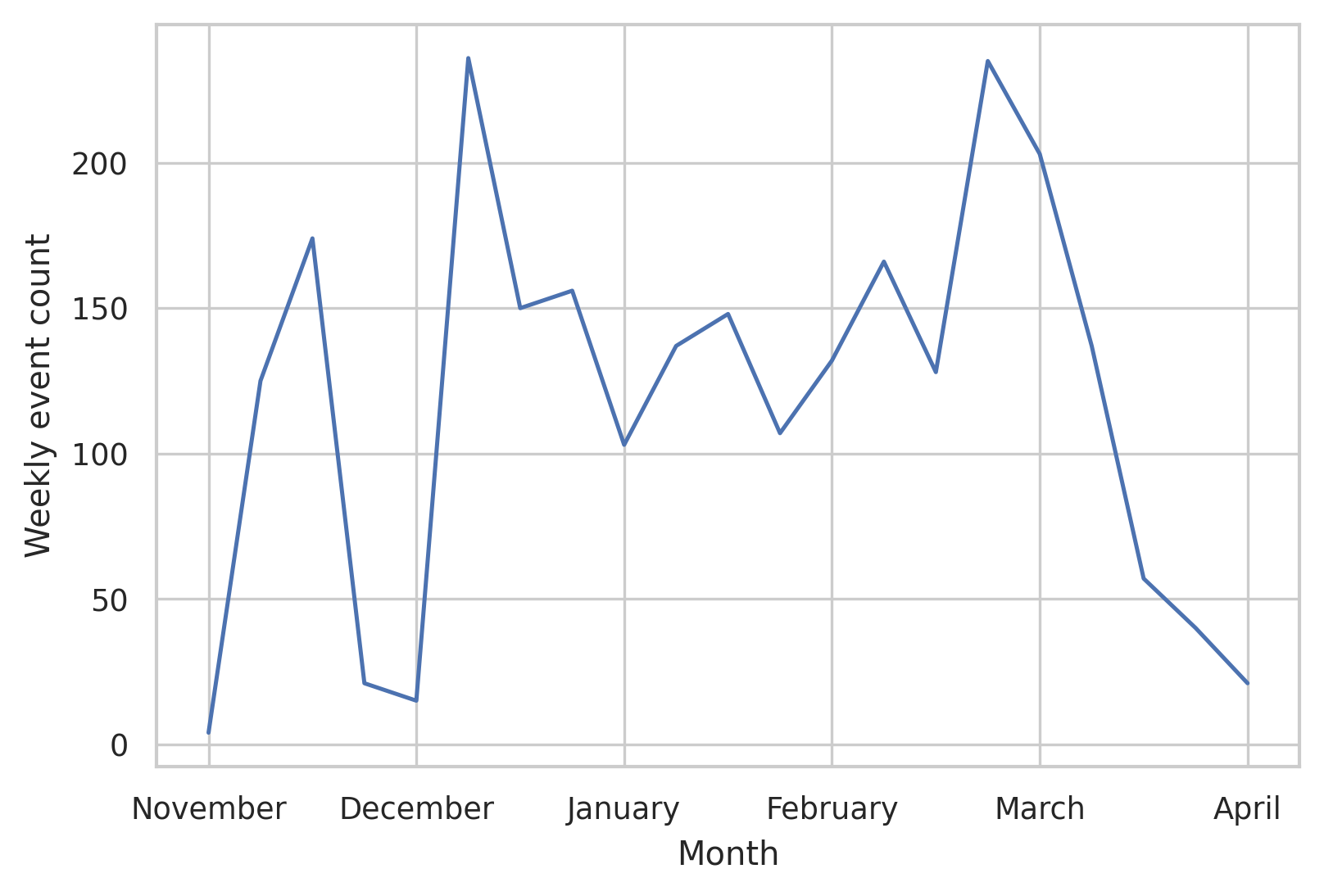}
    \end{minipage}
    \vspace{-0.2in}
    \caption{Left: Fraction of positively identified true events (green bars), and images containing application (blue bars). Error bars reflect 68\% confidence intervals. Right: Weekly application events throughout the season. All years are combined with a sum each week. }
    \label{fig:prevalence_and_time}
\end{figure*}

The results of the sampling are given in Figure~\ref{fig:prevalence_and_time}.  For event detection accuracy, buckets $B_0,B_1,B_2$ and $B_3$ has success fractions of 0.04, 0.20, 0.32, 0.61, respectively. Equation~\ref{eq:estimate}, gives a population estimate of 951 ([700, 1,203] 68\% CI)  events across all facilities during February and March.

\section{Discussion and Implications}
\label{sec:implications}

Our findings have substantial implications for environmental regulation. 

First, we have found that despite its significant risks and presumptive ban in Wisconsin, the prevalence of winter application of manure is high. It is precisely because such activity has historically been difficult to monitor that it has engendered widespread concerns. Our work should also enable greater scientific study about runoff risks to winter land application, given some existing uncertainties in the evidence base around, for instance, nutrient loss~\cite{srinivasan2006manure}

Second, the prevalence of land application does not seem consistent with what is publicly reported about exemptions that facilities can seek from WI DNR. Given the opacity of how such exemptions are administered, there are two potential interpretations.  The first is that the rate of granting exemptions is very high, which would suggest that the formal ban does not functionally restrict land application by CAFOs. If so, this suggests much needed reform, such as the expansion of manure storage requirements in permit terms. The second is that many CAFOs may be engaging in land application in violation of their permits. These would be direct violations of the Clean Water Act. Based on qualitative study of annual spreading reports, it appears that there may be significant under-reporting. From a random sample of more than 30 land application events, only 13 had publicly available spreading reports. Of those, only four had reported the event. 
Either way, our evidence suggests that laxness of enforcement and monitoring has resulted in widespread winter land application, in spite of the nominal presumptive ban in winter months. 

Third, one of our aims is to provide information that can readily be used by environmental interest groups or regulators to prioritize scarce resources. Our outputs may be used to develop facility-level risk scores to prioritize field visits and inspections.  Currently, the volume of inspections is extremely low: major facilities permitted under the Clean Water Act are required to be visited once every two years; minor facilities are required to be visited once every five years; and in fact, many facilities are never visited at all~\cite{clean_water_act2014,rechtschaffen2003enforcing}.  Our model output can hence be exceptionally valuable in risk prioritization.  Our detected land application risk can be combined with readily available information about proximity to surface water, 
size of the CAFO, plant cover, and permit terms to prioritize field visits. 

Fourth, the challenges of WI DNR in providing machine readable information about land application self-reports illustrates a broader challenge across government that impedes innovation and accountability: severe fragmentation of data~\cite{engstrom2020government, gao_epa}. Such fragmentation poses severe challenges to the use of machine learning in areas where the social benefits may be large~\cite{glaze2021artificial}.
Recent policy proposals have focused on public investments for providing improved, secure, and privacy-protecting access and integrating such administrative datasets~\cite{o2018us, ho2021building}, which would enable far more innovation. 

Fifth, methodologically, our study yields interesting implications about the complementarities of human and machine cognition. In training our team to label instances of land application, humans found it exceptionally helpful to be able to examine the time series of images. This makes sense, as it enables humans to determine whether the appearance of a potential land application area was sudden. 
Our early experiments, such as the dual CNN, focused on enabling a network to draw such intertemporal inferences. Yet our experiments revealed the static computer vision models outperformed models that had access to the time series of imagery. This suggests that object detection models can drastically cut down on the domain expertise humans require for detecting the subtleties of land application events (e.g., the streaks in snow that are indicative of sequential truck driving paths). At the same time, it also suggests that more work can be done to incorporate temporal methods into computer vision models. 

\section{Limitations}

We note several limitations to our approach. First, while our study provides the first image dataset of land application events, it took significant time to train a team of five students and staff researchers to recognize such events in satellite imagery. We relied on environmental experts to provide input, but expert input time was necessarily limited. It remains hard to validate such imagery externally, precisely because there are few other sources of ground truth.  As a result, our ground truth imagery may miss some instances of land application, particularly when the application technique may have been distinct from the most common methods,\footnote{Ground injection and spray application via hose, for instance, are common techniques that may have other imagery signatures than tractor-drawn spreading. We believe these techniques may be less prevalent in the winter, but our application will still be valuable for the most common land application method.} or erroneously classify some instances of land application. We piloted potential human validation by sending  environmental interest group volunteers to sites, and will be scaling that in the upcoming winter season to provide more robust external validation. While the dataset may seem small for deep learning, many public sector settings will face such constraints, as physical inspections can be costly. Our application hence demonstrates the utility of transfer learning when labels are necessarily sparse and our approach can also be conceived of as a ``weak supervision'' approach to generate higher fidelity data when combined with in-person validation. 

Second, despite Planet Labs providing near daily imagery, after removing those images which are cropped, obscured by cloud, or have clarity issues, we are left with approximately one image a week. Thus, while our system can indeed act as a real-time monitoring system, its efficacy is dependent on the quality of images received. The need to filter out low quality imagery hence also means that we may have missed a number of land application events. Future work might try to lower the image quality standards so that a wider variety of imagery is useful. That said, the availability of satellite imagery is improving rapidly, both in terms of cadence, spatial resolution, and bands, so our demonstration here is likely to improve going forward.  

Third, our application currently only searches for land application in images centered on CAFOs. This choice was made for a number of reasons. One is that there were constraints on the number of images we could label. As such, we chose to focus on the regions near CAFOs, because such regions are the most likely to contain application, thus yielding the highest ratio of positives per image examined. The other is a practical reason to reduce the absolute number of false positives, since the contemplated use case is to prioritize field and inspection visits.  As emphasized in Section~\ref{sec:prevalence}, this means we are not catching all application events performed by all 330 CAFOs. That said, one of the main constraints for manure disposal lies in transportation costs, so most land application events are expected to be close to the actual CAFO facility (see Figure~\ref{fig:map}).  It would, however, be possible to expand the search radius based on prior knowledge about the economies of waste transportation. 

Fourth, we have used an admittedly simple method to group image-level detections into events. The AUC score of 0.63 for both YOLOv5 and Faster R-CNN suggests that there is room for improvement. While we are currently grouping overlapping bounding boxes across images into an event, more sophisticated methods might take into account the amount of overlap between the boxes and the variance in their scores. In general, a  potential avenue for future research would be to more formally develop time series event detection methods to model the information directly. 
Last, it is worth discussing the potential ethical concerns around the potential for surveillance. We emphasize that our method is a tool for resource allocation to be used in conjunction with humans on the ground. That is, all predictions made the model are validated by humans (which must respect local law, such as  access rights, when checking for application). The system augments, but does not substitute for the procedural protections of environmental enforcement. Further mitigating such concerns are the fact that (a) 3m/pixel satellite imagery is only capable of detecting fairly large-scale events, and (b) CAFO operators are already mandated to report land application events. Thus, there is no legitimate claim for privacy with respect to manure application.

\section{Conclusion}
We have introduced the problem CAFO land-application detection from aerial imagery, a consequential setting at the intersection of regulatory policy and computer vision. 
Our main contribution is a real-time system for the detection of winter land application by CAFOs using high-cadence, medium resolution satellite imagery. This was used to allocate field inspectors to suspected application sites at the of the 2021/22 Winter season, with plans to be used throughout the 2022/23 season. We also demonstrated our system's capacity to provide useful retrospective analysis of application events, and detect outlier facilities which apply well above the average. Further, performing stratified sampling of our model predictions, we provide an estimate of 951 application events across 330 CAFO locations in Wisconsin throughout February and March, 2022. Our results suggest that winter land  application is pervasive in spite of winter restrictions, but that machine learning coupled with satellite imagery could improve accountability in this environmentally important area.

\begin{acks}
We thank Hannah Kim for excellent research assistance, Sarfaraz Alam, Brandon Anderson, Katie Garvey, John Klein, Lynn Henning, Rob Michaels, John Petoskey, Christine Tsang, the Socially Responsible Agriculture Project, and the Environmental Law and Policy Center for input, the Environmental Working Group for data, and the Stanford Institute for Economic Policy Research, Stanford Impact Labs, Schmidt Futures, the Chicago Community Trust, the Chicago Community Foundation, and Sarena Snider (Snider Foundation) for supporting this work.
\end{acks}

\bibliographystyle{ACM-Reference-Format}
\bibliography{main}

\end{document}